\documentclass[conference,a4paper]{APSIPA2020}
\usepackage{multirow}
\usepackage[dvips]{graphicx}
\usepackage{amsmath}
\usepackage[psamsfonts]{amssymb}
\usepackage{amsxtra}
\usepackage{threeparttable}
\usepackage{amsmath,graphicx}
\usepackage{bm}
\usepackage{cite}

\usepackage{multirow}

\begin{document}

\title{End-to-End Automatic Speech Recognition with Deep Mutual Learning}

\author{%
  \authorblockN{%
    Ryo Masumura\authorrefmark{1}, Mana Ihori\authorrefmark{1}, Akihiko Takashima\authorrefmark{1}, Tomohiro Tanaka\authorrefmark{1}, Takanori Ashihara\authorrefmark{1}
}
\authorblockA{%
\authorrefmark{1}
NTT Media Intelligence Laboratories, NTT Corporation, Japan\\
E-mail: ryou.masumura.ba@hco.ntt.co.jp}
}

\maketitle
\thispagestyle{empty}
\begin{abstract}
  This paper is the first study to apply deep mutual learning (DML) to end-to-end ASR models. In DML, multiple models are trained simultaneously and collaboratively by mimicking each other throughout the training process, which helps to attain the global optimum and prevent models from making over-confident predictions. While previous studies applied DML to simple multi-class classification problems, there are no studies that have used it on more complex sequence-to-sequence mapping problems. For this reason, this paper presents a method to apply DML to state-of-the-art Transformer-based end-to-end ASR models. In particular, we propose to combine DML with recent representative training techniques. i.e., label smoothing, scheduled sampling, and SpecAugment, each of which are essential for powerful end-to-end ASR models. We expect that these training techniques work well with DML because DML has complementary characteristics. We experimented with two setups for Japanese ASR tasks: large-scale modeling and compact modeling. We demonstrate that DML improves the ASR performance of both modeling setups compared with conventional learning methods including knowledge distillation. We also show that combining DML with the existing training techniques effectively improves ASR performance.
\end{abstract}
\noindent\textbf{Index Terms}: end-to-end ASR, deep mutual learning, Transformer, scheduled sampling, SpecAugment

\section{Introduction}
In the automatic speech recognition (ASR) field, there has been growing interest in developing end-to-end ASR systems that directly convert input speech into text. While traditional ASR systems have been built from noisy channel formulations using several component models (i.e., an acoustic model, language model, and pronunciation model), end-to-end ASR systems can learn the overall conversion in one step without any intermediate processing.

Modeling methods and training techniques help to achieve powerful end-to-end ASR models. Recent studies have developed modeling methods that include connectionist temporal classification \cite{zweig_icassp2017,audhkhasi_interspeech2017}, a recurrent neural aligner \cite{sak_interspeech2017}, a recurrent neural network (RNN) transducer \cite{rao_asru2017}, and an RNN encoder-decoder \cite{bahdanau_icassp2015,lu_interspeech2015,chen_icassp2016,lu_icassp2016,masumura_icassp2019}. In particular, Transformer-based modeling methods have shown the strongest performance in recent studies \cite{vaswani_nips2017,dong_icassp2018,zhou_interspeecg2018,zhao_icassp2019,li_interspeech2019,masumura_interspeech2020}. In addition, a few effective training techniques are label smoothing \cite{szegedy_cvpr2016}, scheduled sampling \cite{bengio_nips2015}, and SpecAugment \cite{park_is2019,park_icassp2020}. These techniques effectively prevent over-fitting problems caused by maximum likelihood estimation, and combining them can improve end-to-end ASR systems \cite{chiu_interspeech2018}. Furthermore, recent studies have focused on building compact models because computation complexity and memory efficiency must be considered in practice. The most representative technique is knowledge distillation \cite{nips_hinton2015} (i.e., teacher-student learning) that trains compact student models to mimic a pre-trained large-scale teacher model. In fact, knowledge distillation is an effective compact end-to-end ASR modeling technique \cite{huang_interspeech2018,kim_icassp2019,masumura_icassp2020}.

To achieve a more powerful and compact end-to-end ASR model, we focused on deep mutual learning (DML) \cite{zhang_cvpr2018}, one of the most successful learning strategies in recent machine learning studies. In DML, multiple student models simultaneously learn to solve a target task collaboratively without introducing pre-trained teacher models. In fact, each student model is constrained to mimic other student models, thereby helping it to find a global optimum and prevent it from making over-confident predictions. DML enables us to construct stronger models using a unified network structure rather than independent learning. In addition, DML can be used to obtain compact models that perform better than those distilled from a strong but static teacher. In previous studies, DML was used on simple multi-class classification problems, such as image classification \cite{zhang_cvpr2018,zhao_icip2019,wu_cvpr2019,thoker_icip2019}. However, no studies have tried DML on more complex sequence-to-sequence mapping problems.

This paper presents a method to incorporate DML in state-of-the-art Transformer-based end-to-end ASR models. In particular, we propose to combine DML with the existing training techniques for end-to-end ASR models. DML is closely related to label smoothing \cite{szegedy_cvpr2016}; both aim to prevent models from making over-confident predictions. While label smoothing uses a uniform distribution to smooth the ground-truth distribution, DML leverages the distributions predicted by other student models. Combining both kinds of smoothing should efficiently prevent over-confident predictions. In addition, DML is related to scheduled sampling \cite{bengio_nips2015} and SpecAugment \cite{park_is2019,park_icassp2020}. While scheduled sampling and SpecAugment aim to maintain consistency between similar conditioning contexts, DML aims to maintain consistency between different student models. We expect that these consistency strategies complement each other.

Our experiments using the Corpus of Spontaneous Japanese (CSJ) \cite{maekawa_lrec2000} examined two experimental setups: large-scale modeling and compact modeling. We found that DML improves the ASR performance of both modeling setups compared with conventional learning methods, including knowledge distillation. We also found that combining DML with the existing training techniques effectively improves ASR performance.

\section{End-to-End ASR with Transformer}
This section briefly describes end-to-end ASR that uses Transformer-based encoder-decoder models based on auto-regressive generative modeling \cite{vaswani_nips2017,dong_icassp2018,zhou_interspeecg2018,zhao_icassp2019,li_interspeech2019,karita_interspeech2019}. The encoder-decoder models predict a generation probability of a text $\bm{W}=\{w_1,\cdots,w_N\}$ given speech $\bm{X}=\{\bm{x}_1,\cdots,\bm{x}_M\}$, where $w_n$ is the $n$-th token in the text and $\bm{x}_m$ is the $m$-th acoustic feature in the speech. $N$ is the number of tokens in the text and $M$ is the number of acoustic features in the speech. In the auto-regressive generative models, the generation probability of $\bm{W}$ is defined as
\begin{equation}
  P(\bm{W}|\bm{X}; \bm{\Theta}) = \prod_{n=1}^N P(w_n|\bm{W}_{1:n-1}, \bm{X}; \bm{\Theta}) ,
\end{equation}
where $\bm{\Theta}$ represents the trainable model parameter sets and $\bm{W}_{1:n-1}=\{w_1,\cdots,w_{n-1}\}$. In our Transformer-based end-to-end ASR models, $P(w_n|\bm{W}_{1:n-1}$ $\bm{X}$; $\bm{\Theta})$ is computed using a speech encoder and a text decoder, both of which are composed of a couple of Transformer blocks.

\begin{figure*}[t]
  \begin{center}
    \includegraphics[width=130mm]{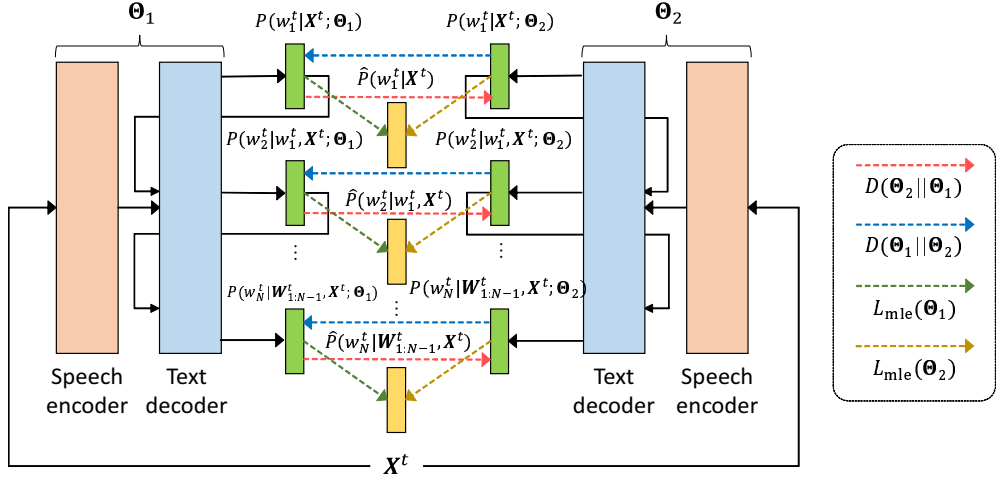}
  \end{center}
  \vspace{-3mm}
  \caption{Deep mutual learning using two student models.}
\end{figure*}

\subsection{Network structure}

\smallskip
\noindent {\bf Speech encoder:}
The speech encoder converts input acoustic features $\bm{X}$ into the hidden representations $\bm{H}^{(I)}$ using $I$ Transformer encoder blocks. The $i$-th Transformer encoder block composes $i$-th hidden representations $\bm{H}^{(i)}$ from the lower layer inputs $\bm{H}^{(i-1)}$ as
\begin{equation}
  \bm{H}^{(i)} = {\tt TransformerEncoderBlock}(\bm{H}^{(i-1)}; \bm{\Theta}) ,
\end{equation}
where ${\tt TransformerEncoderBlock}()$ is a Transformer encoder block that consists of a scaled dot product multi-head self-attention layer and a position-wise feed-forward network \cite{vaswani_nips2017}. The hidden representations $\bm{H}^{(0)}=\{\bm{h}_1^{(0)},\cdots,\bm{h}_{M^{\prime}}^{(0)}\}$ are produced by
\begin{equation}
  \bm{h}_{m^\prime}^{(0)} = {\tt AddPostionalEncoding}(\bm{h}_{m^\prime}) , 
\end{equation}
where ${\tt AddPositionalEncoding}()$ is a function that adds a continuous vector in which position information is embedded. $\bm{H}= \{\bm{h}_1,\cdots,\bm{h}_{M^{\prime}}\}$ is produced by
\begin{equation}
  \bm{H} = {\tt ConvolutionPooling}(\bm{x}_1,\cdots,\bm{x}_{M}; \bm{\Theta}) ,
\end{equation}
where ${\tt ConvolutionPooling}()$ is a function composed of convolution layers and pooling layers. $M^{\prime}$ is the subsampled sequence length depending on the function.

\smallskip
\noindent {\bf Text decoder:}
The text decoder computes the generative probability of a token from preceding tokens and the hidden representations of the speech. The predicted probabilities of the $n$-th token $w_n$ are calculated as
\begin{equation}
  P(w_n|\bm{W}_{1:n-1}, \bm{X}; \bm{\Theta}) = {\tt Softmax}(\bm{u}_{n-1}^{(J)}; \bm{\Theta}) ,
\end{equation}
where ${\tt Softmax}()$ is a softmax layer with a linear transformation. The input hidden vector $\bm{u}_{n-1}^{(J)}$ is computed from $J$ Transformer decoder blocks. The $j$-th Transformer decoder block composes $j$-th hidden representation $\bm{u}_{n-1}^{(j)}$ from the lower layer inputs $\bm{U}_{1:n-1}^{(j-1)} = \{\bm{u}_{1}^{(j-1)},\cdots,\bm{u}_{n-1}^{(j-1)}\}$ as
\begin{equation}
  \bm{u}_{n-1}^{(j)} = {\tt TransformerDecoderBlock}(\bm{U}^{(j-1)}_{1:n-1}, \bm{H}^{(I)}; \bm{\Theta}) ,
\end{equation}
where ${\tt TransformerDecoderBlock}()$ is a Transformer decoder block that consists of a scaled dot product multi-head masked self-attention layer, a scaled dot product multi-head source-target attention layer, and a position-wise feed-forward network \cite{vaswani_nips2017}. The hidden representations $\bm{U}_{1:n-1}^{(0)}=\{\bm{u}_1^{(0)},\cdots,\bm{u}_{n-1}^{(0)}\}$ are produced by
\begin{equation}
  \bm{u}_{n-1}^{(0)} =  {\tt AddPositionalEncoding}(\bm{w}_{n-1}) ,
\end{equation}
\begin{equation}
  \bm{w}_{n-1} = {\tt Embedding}(w_{n-1}; \bm{\Theta}) , 
\end{equation}
where ${\tt Embedding}()$ is a linear layer that embeds input token in a continuous vector.

\subsection{Typical objective function}
In end-to-end ASR, a model parameter set can be optimized from the utterance-level training data set ${\cal U} =$ $\{(\bm{X}^1, \bm{W}^1),$ $\cdots,$ $(\bm{X}^{T},$ $\bm{W}^{T})\}$, where $T$ is the number of utterances in the training data set. An objective function based on the maximum likelihood estimation is defined as
\begin{multline}
  {\cal L}_{\tt mle}(\bm{\Theta}) = - \sum_{t = 1}^{T} \sum_{n=1}^{N^t} \sum_{w^t_n \in{\cal V}} {\hat P}(w^t_n|\bm{W}_{1:n-1}^t, \bm{X}^t) \\
  \log P(w^t_n|\bm{W}_{1:n-1}^t, \bm{X}^t; \bm{\Theta}) ,
\end{multline}
where $w_n^t$ is the $n$-th token for the $t$-th utterance and $\bm{W}_{1:n-1}^t = \{w_1^t,\cdots,w_{n-1}^t\}$. $\cal V$ represents the vocabulary sets, and $N^t$ is the number of tokens in the $t$-th utterance. The ground-truth probability ${\hat P}(w_n^t|\bm{W}_{1:n-1}^t, \bm{X}^t)$ is defined as
\begin{equation}
{\hat P}(w_n^t|\bm{W}_{1:n-1}^t, \bm{X}^t) = 
\begin{cases}
  1 & (w_n^t = \hat{w}_n^t)  \\
  0 & (w_n^t \neq \hat{w}_n^t) ,
\end{cases}
\end{equation}
where $\hat{w}_n^t$ is the $n$-th reference token in the $t$-th utterance.

\subsection{Training techniques}
There are several training techniques in the end-to-end ASR modeling. This paper introduces the following three techniques.

\smallskip
\noindent {\bf Label smoothing:}
Label smoothing is a regularization technique that can prevent the model from making over-confident predictions \cite{szegedy_cvpr2016}. This encourages the model to have higher entropy at its prediction. This paper introduces a uniform distribution to all tokens in vocabulary that smooths the ground-truth probabilities. Thus, an objective function that uses the label smoothing is defined as
\begin{multline}
  {\cal L}_{\tt ls}(\bm{\Theta}) = - \sum_{t = 1}^{T} \sum_{n=1}^{N^k} \sum_{w_n^t \in{\cal V}}  {\tilde P}(w_n^t|\bm{W}_{1:n-1}^t, \bm{X}^t)  \\
  \log P(w_n^t|\bm{W}_{1:n-1}^t, \bm{X}^t; \bm{\Theta}) ,
\end{multline}
 \vspace{-6mm}
\begin{multline}
  {\tilde P}(w_n^t|w_1^t,\cdots,w_{n-1}^t, \bm{X}^t) = \\
  (1-\alpha) {\hat P}(w_n^t|\bm{W}_{1:n-1}^t, \bm{X}^t) + \alpha \frac{1}{|\cal V|} ,
\end{multline}
where $\alpha$ is a smoothing weight to adjust the smoothing term.

\smallskip
\noindent {\bf Scheduled sampling:}
Scheduled sampling is a technique that randomly uses predicted tokens as conditioning tokens in the text decoder \cite{bengio_nips2015}. This technique helps reduce the gap between teacher forcing in a training phase and free running in a testing phase. An objective function that uses scheduled sampling is defined as
\begin{multline}
  {\cal L}_{\tt ss}(\bm{\Theta}) = - \sum_{t = 1}^{T} \sum_{n=1}^{N^k} \sum_{w_n^t \in{\cal V}}  {\hat P}(w_n^t|\bm{W}_{1:n-1}^t, \bm{X}^t)  \\
  \log P(w_n^t|{\cal S}(\bm{W}_{1:n-1}^t), \bm{X}^t; \bm{\Theta}) ,
\end{multline}
where ${\cal S}()$ is a scheduled sampling function with random behavior for the conditioning tokens.

\smallskip
\noindent {\bf SpecAugment:}
SpecAugment is a technique that augments input acoustic feature representations \cite{park_is2019,park_icassp2020}. This technique consists of three kinds of deformations: time warping, time masking, and frequency masking. Time warping is a deformation of the acoustic features in the time direction. Time masking and frequency masking mask a block of consecutive time steps or frequency channels. An objective function that uses SpecAugment is defined as
\begin{multline}
  {\cal L}_{\tt sa}(\bm{\Theta}) = - \sum_{t = 1}^{T} \sum_{n=1}^{N^k} \sum_{w_n^t \in{\cal V}}  {\hat P}(w_n^t|\bm{W}_{1:n-1}^t, \bm{X}^t)  \\
  \log P(w_n^t|\bm{W}_{1:n-1}^t, {\cal G}(\bm{X}^t); \bm{\Theta}) ,
\end{multline}
where ${\cal G}()$ is the SpecAugment deformation function with random behavior for the input acoustic features.

\section{Proposed Method}
This section details deep mutual learning (DML) for end-to-end ASR. In addition, we present objective functions when combining DML with several training techniques.

\subsection{Deep mutual learning for end-to-end ASR}
In DML, $K$ different model parameters $\{\bm{\Theta}_1,\cdots,\bm{\Theta}_K\}$ are simultaneously trained to mimic each other, while the conventional training method learns the model parameters to predict ground-truth probabilities for the training instances. Figure 1 represents DML using two student model parameters. A DML-based objective function for training the $k$-th model parameter $\bm{\Theta}_k$ is defined as
\begin{equation}
  {\cal L}_{\tt dml}(\bm{\Theta}_k) = (1-\lambda) {\cal L}_{\tt mle}(\bm{\Theta}_k) +  \lambda  \frac{1}{K-1} \sum_{i=1, i \neq k}^K  {\cal D}(\bm{\Theta}_i||\bm{\Theta}_k) ,
\end{equation}
where $D(\bm{\Theta}_i||\bm{\Theta}_k)$ is a mimicry loss to mimic the $i$-th model, and $\lambda$ is an interpolation weight to adjust the influence of the mimicry loss. The mimicry loss is computed from
\begin{multline}
  {\cal D}(\bm{\Theta}_i||\bm{\Theta}_k) = - \sum_{t = 1}^{T} \sum_{n=1}^{N^k} \sum_{w_n^t \in {\cal V}} P(w_n^t|\bm{W}_{1:n-1}^t,\bm{X}^t; \bm{\Theta}_i) \\
  \log P(w_n^t|\bm{W}_{1:n-1}^t, \bm{X}^t; \bm{\Theta}_k) .
\end{multline}

In a mini-batch training, $K$ model parameters are optimized jointly and collaboratively.
Thus, $K$ models are learned with the same mini-batches. In each mini-batch step, we compute predicted probability distributions using the $K$ models and update each parameter according to the predicted probability distributions of the others. These optimizations are conducted iteratively until convergence. We finally pick up the single model with the smallest validation loss or a pre-defined compact model.

\subsection{Deep mutual learning with training techniques}
DML can be combined with existing training techniques for end-to-end ASR. This paper proposes new objective functions specific to using DML with label smoothing, scheduled sampling, and SpecAugment. Note that all techniques can be simultaneously combined with DML.

\smallskip
\noindent {\bf Deep mutual learning with label smoothing:}
Both label smoothing and DML avoid peaky predictions with very low entropy. When combining label smoothing with DML, we define an objective function that trains $k$-th model parameter $\bm{\Theta}_k$ as
\begin{equation}
  {\cal L}_{\tt dml+ls}(\bm{\Theta}_k) = (1-\lambda) {\cal L}_{\tt ls}(\bm{\Theta}_k) + \lambda  \frac{1}{K-1} \sum_{i=1, i \neq k}^K  {\cal D}(\bm{\Theta}_i||\bm{\Theta}_k) ,
\end{equation}
where ${\cal L}_{\tt ls}$ is the same as Eq. (11).

\smallskip
\noindent {\bf Deep mutual learning with scheduled sampling:}
When combining scheduled sampling with DML, we aim to make model more robust to various conditioning tokens by maintaining consistency between different models with different conditioning contexts. Thus, an objective function for the $k$-th model parameter is defined as
\begin{multline}
  {\cal L}_{\tt dml+ss}(\bm{\Theta}_k) = (1-\lambda) {\cal L}_{\tt ss}(\bm{\Theta}_k) \\
  +  \lambda  \frac{1}{K-1} \sum_{i=1,i \neq k}^K  {\cal D}_{\tt ss}(\bm{\Theta}_i||\bm{\Theta}_k) ,
\end{multline}
 \vspace{-6mm}
\begin{multline}
  {\cal D}_{\tt ss}(\bm{\Theta}_i||\bm{\Theta}_k) = 
  - \sum_{t = 1}^{T} \sum_{n=1}^{N^k} \sum_{w_n^t \in {\cal V}} \\
  P(w_n^t|{\cal S}_i(\bm{W}_{1:n-1}^t),\bm{X}^t; \bm{\Theta}_i) \\
  \log P(w_n^t|{\cal S}_k(\bm{W}_{1:n-1}^t), \bm{X}^t; \bm{\Theta}_k) ,
\end{multline}
where ${\cal L}_{\tt ss}$ is the same as Eq. (13).
${\cal S}_i()$ and ${\cal S}_k()$ are the functions for the scheduled sampling with different random seeds.

\smallskip
\noindent {\bf Deep mutual learning with SpecAugment:}
When combining SpecAugment with DML, we aim to make model more robust to various acoustic feature examples by maintaining consistency between different models with different deformation. An objective function for the $k$-th model parameter is defined as
\begin{multline}
  {\cal L}_{\tt dml+sa}(\bm{\Theta}_k) = (1-\lambda) {\cal L}_{\tt sa}(\bm{\Theta}_k) \\
  + \lambda  \frac{1}{K-1} \sum_{i=1,i \neq k}^K  {\cal D}_{\tt sa}(\bm{\Theta}_i||\bm{\Theta}_k) ,
\end{multline}
 \vspace{-6mm}
\begin{multline}
  {\cal D}_{\tt sa}(\bm{\Theta}_i||\bm{\Theta}_k) = - \sum_{t = 1}^{T} \sum_{n=1}^{N^k} \sum_{w_n^t \in {\cal V}} \\
  P(w_n^t|\bm{W}_{1:n-1}^t,{\cal G}_i(\bm{X}^t); \bm{\Theta}_i) \\
  \log P(w_n^t|\bm{W}_{1:n-1}^t, {\cal G}_k(\bm{X}^t); \bm{\Theta}_k) ,
\end{multline}
where ${\cal L}_{\tt sa}$ is the same as Eq. (14). ${\cal G}_i()$ and ${\cal G}_k()$ are the functions for the SpecAugment with different random seeds.


\setcounter{table}{0}
\begin{table*}[t!]
  \begin{center}
      \caption{Experimental results in terms of character error rate (\%).}
    \begin{tabular}{|l|ccccc|rrr|} \hline
      & Label & Scheduled & \multirow{2}{*}{SpecAugment} & Knowledge & Deep mutual learning & \multirow{2}{*}{Test 1} & \multirow{2}{*}{Test 2} & \multirow{2}{*}{Test 3} \\
      & smoothing & sampling &  & distillation & (DML) & & & \\  \hline \hline
      Large-scale & - & - & - & - & - & 8.83 & 6.49 & 7.19 \\
      modeling &  $\surd$ & - & - & - & - & 8.41 & 6.23 & 6.74 \\
      & - & $\surd$ & - & - & - & 8.59 & 6.31 & 6.30 \\
      & - & - &  $\surd$ & - & - &  7.48 & 5.59 & 5.86 \\
      &  $\surd$ &  $\surd$ & $\surd$ & - & - & 7.24 & 5.13 & 5.40 \\
      & - & - & - & - &  $\surd$ & 8.19 & 5.78 & 6.37 \\
      &  $\surd$ & - & - & - &  $\surd$ & 8.05 & 5.67 & 6.30 \\
      & - & $\surd$ & - & - &  $\surd$ & 7.90 & 5.57 & 5.62 \\
      & - & - &  $\surd$ & - &  $\surd$ & 7.02 & 4.92 & 5.28 \\
      &  $\surd$ & $\surd$ &  $\surd$ & - &  $\surd$ & {\bf 6.87} & {\bf 4.73} & {\bf 5.02} \\ \hline \hline
      Compact & - & - & - & - & - & 12.80 & 9.43 & 10.01 \\
      modeling &  $\surd$ &  $\surd$ &  $\surd$ & - & - & 11.37 & 7.88 & 8.44 \\
      & - & - & - &  $\surd$ & - & 11.67 & 8.28 & 9.08 \\
      &  $\surd$ &  $\surd$ &  $\surd$ & $\surd$ & - & 11.15 & 7.58 & 8.31 \\  
      & - & - & - & - &  $\surd$ & 11.23 & 7.82 & 8.74 \\
      &  $\surd$ &  $\surd$ &  $\surd$ & - &  $\surd$ & {\bf 10.65} & {\bf 7.19} & {\bf 7.93} \\  \hline
    \end{tabular}
  \end{center}
\end{table*}

\section{Experiments}
We experimented using CSJ \cite{maekawa_lrec2000}. We divided the CSJ into a training set (512.6 hours), a validation set (4.8 hours), and three test sets (1.8 hours, 1.9 hours, and 1.3 hours). We used the validation set to choose several hyper parameters and to conduct early stopping. Each discourse-level speech was segmented into utterances in accordance with our previous work \cite{masumura_interspeech2019}. We used characters as the tokens.

\subsection{Setups}
We examined two types of experimental setups: large-scale modeling and compact modeling.
\begin{itemize}
\item {\bf Large-scale modeling}: We set $I=8$ for the encoder blocks and $J=6$ for the decoder blocks. When introducing DML, we prepared 4 large-scale models and evaluated the single model with the least validation loss. 
\item {\bf Compact modeling}: We set $I=2$ for the encoder blocks and $J=1$ for the decoder blocks where other parameters were the same as the large-scale modeling. When introducing the knowledge distillation \cite{nips_hinton2015} or the deep mutual learning, we prepared 1 compact model and 3 large-scale models and evaluated the compact model. 
\end{itemize}

In both setups, Transformer blocks were composed using the following conditions: the dimensions of the output continuous representations were set to 256, the dimensions of the inner outputs in the position-wise feed forward networks were set to 2,048, and the number of heads in the multi-head attentions was set to 4. For the speech encoder, we used 40 log mel-scale filterbank coefficients appended with delta and acceleration coefficients as acoustic features. The frame shift was 10 ms. The acoustic features passed two convolution and max pooling layers with a stride of 2, so we downsampled them to $1/4$ along with the time-axis. In the text decoder, we used 256-dimensional word embeddings. We set the vocabulary size to 3,262.

For the optimization, we used the Adam optimizer with $\beta_1 = 0.9$, $\beta_2 = 0.98$, $\epsilon = 10^{-9}$ and varied the learning rate based on the update rule presented in previous studies \cite{vaswani_nips2017}. The training steps were stopped based on early stopping using the validation set. We set the mini-batch size to 32 utterances and the dropout rate in the Transformer blocks to 0.1. When we introduced label smoothing, we set $\alpha$ as 0.1. Our scheduled sampling-based optimization process used the teacher forcing at the beginning of the training steps, and we linearly ramped up the probability of sampling to the specified probability at the specified epoch (20 epoch). Our SpecAugment only applied frequency masking and time masking where the number of frequency masks and time step masks were set to 2, the frequency masking width was randomly chosen from 0 to 20 frequency bins, and the time masking width was randomly chosen from 0 to 100 frames. $\lambda$ was set to 0.4 in DML. We used a beam search algorithm in which the beam size was set to 20.

\subsection{Results}
We evaluated various setups using DML and the training techniques in large-scale modeling and compact modeling setups. Table 1 shows experimental results in terms of character error rate. 

First, in the large-scale modeling setup, the results show that each training tip improves Transformer-based end-to-end ASR performance, and combining the techniques effectively improved ASR performance. SpecAugment significantly improved performance in particular. These results indicate that training techniques are important for the Transformer-based end-to-end ASR models. In addition, we improved ASR performance by introducing DML into the Transformer-based end-to-end ASR models, both with and without training techniques. It is thought that DML could help discover the global optimum and prevent models from making over-confident predictions. The highest results were attained by combining DML and all the training techniques. This suggests that combining DML with existing training techniques effectively improves ASR performance.

Next, in the compact modeling setups, the results show that DML improved performance even more than knowledge distillation. This indicates that DML in which student models interact with each other during all the training steps effectively transfers knowledge in large-scale end-to-end ASR models to the compact end-to-end ASR models. These results confirm that DML is a good solution to build Transformer-based end-to-end ASR models.

\section{Conclusions}
We have presented a method to incorporate deep mutual learning (DML) in Transformer-based end-to-end automatic speech recognition models. The key advance of our method is to introduce combined training strategies of DML with representative training techniques (label smoothing, scheduled sampling, and SpecAugment) for end-to-end ASR models. Our experiments demonstrated that the DML improves ASR performance of both large-scale modeling and compact modeling setups compared with conventional learning methods, including knowledge distillation. We also showed that combining DML with existing training techniques effectively improves ASR performance.

\end{document}